\documentclass[11pt]{article}
\usepackage{eacl2017}
\usepackage{times}
\usepackage{url}
\usepackage{latexsym}
\usepackage{booktabs}
\usepackage{color}
\usepackage{enumitem}
\usepackage{graphicx}
\usepackage[T1]{fontenc}
\usepackage{caption}
\usepackage{algorithm}
\usepackage{graphics}
\usepackage{algorithmic}
\usepackage[utf8]{inputenc}
\usepackage{url}
\usepackage{tikz}
\usepackage{tikz-dependency}

\eaclfinalcopy 
\def\eaclpaperid{281} 

\title{Parsing Universal Dependencies without training}

\author{Héctor Martínez Alonso$^\spadesuit$ \quad Željko Agić$^\heartsuit$ \quad Barbara Plank$^\clubsuit$ \quad Anders Søgaard$^\diamondsuit$\\[0.5ex]
$^\spadesuit$Univ. Paris Diderot, Sorbonne Paris Cité -- Alpage, INRIA, France\\
$^\heartsuit$IT University of Copenhagen, Denmark\\
$^\clubsuit$Center for Language and Cognition, University of Groningen, The Netherlands\\
$^\diamondsuit$University of Copenhagen, Denmark\\[0.5ex]
{\tt hector.martinez-alonso@inria.fr}}

\date{}

\begin{document}
\maketitle
\begin{abstract}
We propose UDP, the first training-free parser for Universal Dependencies (UD). Our algorithm is based on 
PageRank and a small set of head attachment rules. It features two-step decoding to guarantee that function words are attached as leaf nodes. The parser requires no training, and it is competitive with a delexicalized transfer system. UDP offers a linguistically sound unsupervised alternative to cross-lingual parsing for UD, which can be used as a baseline for such systems. The parser has very few parameters and is distinctly robust to domain change across languages.
\end{abstract}
\section{Introduction}

Grammar induction and unsupervised dependency parsing are active fields of research in natural language processing \cite{klein2004corpus,gelling2012pascal}. However, many data-driven approaches struggle with learning relations that match the conventions of the test data, e.g., Klein and Manning reported the tendency of their DMV parser to make determiners the heads of German nouns, which would not be an error if the test data used a DP analysis~\cite{abney1987english}.
Even supervised transfer approaches \cite{mcdonald2011multi} suffer from target adaptation problems when facing word order differences.

The Universal Dependencies (UD) project \cite{UD12,nivre2016universal} offers a dependency formalism that aims at providing a consistent representation across languages, while enforcing a few hard constraints. The arrival of such treebanks, expanded and improved on a regular basis, provides a new milestone for cross-lingual dependency parsing research \cite{mcdonald2013universal}. 

\noindent
Furthermore, given that UD rests on a series of simple principles like the primacy of lexical heads, cf. \newcite{johannsen2015universal} for more details, we expect that such a formalism lends itself more naturally to a simple and linguistically sound rule-based approach to cross-lingual parsing. In this paper we present such an approach.

Our system is a dependency parser that requires no training, and relies solely on explicit part-of-speech (POS) constraints that UD imposes. In particular, UD prescribes that trees are single-rooted, and that function words like adpositions, auxiliaries, and determiners are always dependents of content words, while other formalisms might treat them as heads \cite{marneffe2014universal}. We ascribe our work to the viewpoints of \newcite{Bender2009naive} about the incorporation of linguistic knowledge in language-independent systems.

\paragraph{Contributions } We introduce, to the best of our knowledge, the first unsupervised rule-based dependency parser for Universal Dependencies.

Our method goes substantially beyond the existing work on \emph{rule-aided} unsupervised dependency parsing, specifically by:
\begin{itemize}[noitemsep,topsep=3pt]
\item[i)] adapting the dependency head rules to UD-compliant POS relations, 
\item[ii)] incorporating the UD restriction of function words being leaves, 
\item[iii)] applying personalized PageRank to improve main predicate identification, and by
\item[iv)] making the parsing entirely free of language-specific parameters by estimating adposition attachment direction at runtime.
\end{itemize}
We evaluate our system on 32 languages\footnote{Out of 33 languages in UD v1.2. We exclude Japanese because the treebank is distributed without word forms and hence we can not provide results on predicted POS.} in three setups, depending on the reliability of available POS tags, and compare to a multi-source delexicalized transfer system. 
In addition, we evaluate the systems' sensitivity to domain change for a subset of UD languages for which domain information was retrievable. The results expose a solid and competitive system for all UD languages. Our unsupervised parser compares favorably to delexicalized parsing, while being more robust to domain change.

\section{Related work}
\label{sec:relatedwork}

\paragraph{Cross-lingual learning} Recent years have seen exciting developments in cross-lingual linguistic structure prediction based on transfer or projection of POS and dependencies \cite{das2011unsupervised,mcdonald2011multi}. These works mainly use supervised learning and domain adaptation techniques for the target language. 

The first group of approaches deals with annotation projection \cite{yarowsky2001inducing}, whereby parallel corpora are used to transfer annotations between resource-rich source languages and low-resource target languages. Projection relies on the availability and quality of parallel corpora, source-side taggers and parsers, but also tokenizers, sentence aligners, and word aligners for sources and targets. \newcite{hwa2005bootstrapping} were the first to project syntactic dependencies, and Tiedemann et al.~\shortcite{tiedemann2014rediscovering,tiedemann2016synthetic} improved on their projection algorithm. Current state of the art in cross-lingual dependency parsing involves leveraging parallel corpora for annotation projection \cite{ma2014unsupervised,rasooli2015density}.

The second group of approaches deals with transferring source parsing models to target languages. \newcite{zeman2008cross} were the first to introduce the idea of delexicalization: removing lexical features by training and cross-lingually applying parsers solely on POS sequences. \newcite{sogaard2011data} and \newcite{mcdonald2011multi} independently extended the approach by using multiple sources, requiring uniform POS and dependency representations \cite{mcdonald2013universal}.

Both model transfer and annotation projection rely on a large number of presumptions to derive their competitive parsing models. By and large, these presumptions are unrealistic and exclusive to a group of very closely related, resource-rich Indo-European languages. Agić et al.~\shortcite{agic2015if,agic2016multilingual} exposed some of these biases in their proposal for realistic cross-lingual tagging and parsing, as they emphasized the lack of perfect sentence- and word-splitting for truly low-resource languages. Further, \newcite{johannsen2016joint} introduced joint projection of POS and dependencies from multiple sources while sharing the outlook on bias removal in real-world multilingual processing.

\paragraph{Rule-based parsing} Cross-lingual methods, realistic or not, depend entirely on the availability of data: for the sources, for the targets, or most often for both sets of languages. Moreover, they typically do not exploit constraints placed on linguistic structures through a formalism, and they do so {\em by design}.

With the emergence of UD as the practical standard for multilingual POS and syntactic dependency annotation, we argue for an approach that takes a fresh angle on both aspects. Specifically, we propose a parser that i) requires {\em no} training data, and in contrast ii) critically {\em relies} on exploiting the UD constraints.

These two characteristics make our parser unsupervised. Data-driven unsupervised dependency parsing is now a well-established discipline \cite{klein2004corpus,spitkovsky2010baby,spitkovsky2010viterbi}. Still, the performance of these parsers falls far behind the approaches involving any sort of supervision.

Our work builds on the line of research on rule-aided unsupervised dependency parsing by \newcite{gillenwater2010sparsity} and \newcite{naseem2010using}, and also relates to Søgaard's~\shortcite{sogaard2012two,sogaard2012unsupervised} work. Our parser, however, features two key differences:
\begin{itemize}[noitemsep,topsep=3pt]
\item[ i)] the usage of PageRank personalization \cite{lofgren2015efficient}, and of 
\item[ii)] two-step decoding to treat content and function words differently according to the UD formalism. 
\end{itemize}
Through these differences, even without any training data, we parse nearly as well as a delexicalized transfer parser, and with increased stability to domain change.

\section{Method}
\label{sec:method}
Our approach does not use any training or unlabeled data. We have used the English treebank during development to assess the contribution of individual head rules, and to tune PageRank parameters (Sec.\ \ref{sec:pagerank}) and function-word directionality (Sec.\ \ref{sec:directionality}). Adposition direction is calculated on the fly at runtime. We refer henceforth to our UD parser as UDP. 

\subsection{PageRank setup}
\label{sec:pagerank}
Our system uses the PageRank (PR) algorithm \cite{page1999pagerank} to estimate the relevance of the content words of a sentence. PR uses a random walk to estimate which nodes in the graph are more likely to be visited often, and thus, it gives higher rank to nodes with more incoming edges, as well as to nodes connected to those. Using PR to score word relevance requires an effective graph-building strategy. We have experimented with the strategies by \newcite{sogaard2012unsupervised}, such as words being connected to adjacent words, but our system fares best strictly using the dependency rules in Table~\ref{tab:rules} to build the graph. UD trees are often very flat, and a highly connected graph yields a PR distribution that is closer to uniform, thereby removing some of the difference of word relevance.

We build a multigraph of all words in the sentence covered by the head-dependent rules in Table~\ref{tab:rules}, giving each word an incoming edge for each eligible dependent, i.e., \textsc{adv} depends on \textsc{adj} and \textsc{verb}. This strategy does not always yield connected graphs, and we use a teleport probability of 0.05 to ensure PR convergence. 

\textit{Teleport probability} is the probability that, in any iteration of the PR calculation, the next active node is randomly chosen, instead of being one of the adjacent nodes of the current active node. See \newcite{brin2012reprint} for more details on teleport probability, where the authors refer to one minus teleport probability as \textit{damping factor}.

We chose this value incrementally in intervals of 0.01 during development until we found the smallest value that guaranteed PR convergence. A high teleport probability is undesirable, because the resulting stationary distribution can be almost uniform. We did not have to re-adjust this value when running on the actual test data.

The main idea behind our personalized PR approach is the observation that ranking is only relevant for content words.\footnote{\textsc{adj}, \textsc{noun}, \textsc{propn}, and \textsc{verb} mark content words.} PR can incorporate a priori knowledge of the relevance of nodes by means of \emph{personalization}, namely giving more weight to certain nodes. 

Intuitively, the higher the rank of a word, the closer it should be to the root node, i.e., the main predicate of the sentence is the node that should have the highest PR, making it the dependent of the root node (Fig.~\ref{fig:alg}, lines 4-5). We use PR personalization to give 5 times more weight (over an otherwise uniform distribution) to the node that is estimated to be main predicate, i.e., the first verb or the first content word if there are no verbs.

\subsection{Head direction} 
\label{sec:directionality}

Head direction is an important trait in dependency syntax~\cite{tesniere1959elements}. Indeed, the UD feature inventory contains a trait to distinguish the general adposition tag \textsc{adp} in pre- and post-positions. 

Instead of relying on this feature from the treebanks, which is not always provided, we estimate the frequency of \textsc{adp-nominal} vs. \textsc{nominal-adp} bigrams.\footnote{\textsc{nominal}$=\{$\textsc{noun, propn, pron}$\}$} We calculate this estimation directly on input data at runtime to keep the system training-free. Moreover, it requires very few examples to converge (10-15 sentences). If a language has more \textsc{adp-nominal} bigrams, we consider all its \textsc{adp} to be prepositions (and thus dependent of elements at their right). Otherwise, we consider them postpositions. 

For other function words, we have determined on the English dev data whether to make them strictly right- or left-attaching, or to allow either direction. There, \textsc{aux}, \textsc{det}, and \textsc{sconj} are right-attaching, while \textsc{conj} and \textsc{punct} are left-attaching. There are no direction constraints for the rest. Punctuation is a common source of parsing errors that has very little interest in this setup. While we do evaluate on all tokens including punctuation, we also apply a heuristic for the last token in a sentence; if it is a punctuation, we make it a dependent of the main predicate.

\begin{figure}
    \centering
    \small
		\begin{algorithmic}[1]
	\STATE $H = \emptyset$; $D = \emptyset$
	\STATE  $C = \langle c_1, ... c_m\rangle$; $F = \langle f_1, ... f_m\rangle$
	 \FOR{$c \in C$}
	 \IF{$\left\vert{H}\right\vert = 0$}
	 \STATE $h= root$
	 \ELSE
	\STATE $h=$argmin$_{j \in H}$ $ \{ \gamma(j,c)\mid\delta(j,c) \wedge \kappa(j,c) \} $
	 \ENDIF
	 \STATE $H = H \cup \{c\}$
	 \STATE $D = D \cup \{(h,c)\}$
 	\ENDFOR
	 \FOR{$f \in F$}
	 \STATE $h=$argmin$_{j \in H}$ $ \{ \gamma(j,f)\mid\delta(j,f) \wedge \kappa(j,f) \} $
	  \STATE $D = D \cup \{(h,f)\}$
	 \ENDFOR
	 \RETURN $D$
	\end{algorithmic}
    \caption{Two-step decoding algorithm for UDP.}
    \label{fig:alg}
\end{figure}

\begin{figure}
        \centering
          \begin{tabular}{rll}
 		\toprule
 		ADJ & $\longrightarrow$ & ADV \\
 		NOUN & $\longrightarrow$ & ADJ, NOUN, PROPN\\  
        NOUN & $\longrightarrow$ & ADP, DET, NUM\\  
        PROPN & $\longrightarrow$ & ADJ, NOUN, PROPN\\  
        PROPN & $\longrightarrow$ & ADP, DET, NUM\\  
        VERB & $\longrightarrow$ & ADV, AUX, NOUN \\
 		VERB & $\longrightarrow$ & PROPN, PRON, SCONJ \\
 		\bottomrule
		\end{tabular}
        \captionof{table}{UD dependency rules.}
        \label{tab:rules}
\end{figure}
\subsection{Decoding}
\label{sec:decoding}

Fig.~\ref{fig:alg} shows the tree-decoding algorithm. It has two blocks, namely a first block (3-11) where we assign the head of content words according to their PageRank and the constraints of the dependency rules, and a second block (12-15) where we assign the head of function words according to their proximity, direction of attachment, and dependency rules. The algorithm requires:

\begin{enumerate}[noitemsep,topsep=3pt]
\item The PR-sorted list of content words $C$.
\item The set of function words $F$, sorting is irrelevant because function-head assignations are inter-independent.
\item A set $H$ for the current possible heads, and a set $D$ for the dependencies assigned at each iteration, which we represent as head-dependent tuples $(h,d)$.
\item A symbol $root$ for the root node.
\item A function $\gamma(n,m)$ that gives the linear distance between two nodes.
\item A function $\kappa(h,d)$ that returns whether the dependency $(h,d)$ has a valid attachment direction given the POS of the $d$ (cf. Sec.~\ref{sec:directionality}).
\item A function $\delta(h,d)$ that determines whether $(h,d)$ is licensed by the rules in Table~\ref{tab:rules}.
\end{enumerate}
The head assignations in lines 7 and 13 read as follow: the head $h$ of a word (either $c$ or $f$) is the closest element of the current list of heads ($H$) that has the right direction ($\kappa$) and respects the POS-dependency rules ($\delta$). These assignations have a back-off option to ensure the final $D$ is a tree. If the conditions determined by $\kappa$ and $\delta$ are too strict, i.e., if the set of possible heads is empty, we drop the $\delta$ head-rule constraint and recalculate the closest possible head that respects the directionality imposed by $\kappa$. If the set is empty again, we drop both constraints and assign the closest head.

Lines 4 and 5 enforce the single-root constraint. To enforce the leaf status of function nodes, the algorithm first attaches all content words ($C$), and then all function words ($F$) in the second block where H is not updated, thereby ensuring leafness for all $f \in F$. The order of head attachment is not monotonic wrt. PR between the first and second block, and can yield non-projectivities. Nevertheless, it still is a one-pass algorithm. Decoding runs in less than $O(n^2)$, namely $O(n\times\left\vert{C}\right\vert)$. However, running PR incurs the main computation cost.

\section{Parser run example}

This section exemplifies a full run of UDP for the example sentence from the English test data: ``They also had a special connection to some extremists''.
\subsection{PageRank}
Given an input sentence and its POS tags, we obtain rank of each word by building a graph using head rules and running PR on it. Table \ref{tbl:sentence} provides the sentence, the POS of each word, the number of incoming edges for each word after building the graph with the head rules from Sec. \ref{sec:pagerank}, and the personalization vector for PR on this sentence. Note that all nodes have the same personalization weight, except the estimated main predicate, the verb ``had''.

\begin{figure}[ht]
	\resizebox{\columnwidth}{!}{
    \begin{tabular}{ r l l l l l l l l l }
    \toprule
    Word: &They & also & had & a & special & connection & to & some & extremists \\
    POS: &PRON & ADV & VERB & DET & ADJ & NOUN & ADP & DET & NOUN \\
    \midrule
    Personalization:& 1 & 1 & 5 & 1 & 1 & 1 & 1 & 1 & 1 \\
    Incoming edges:&0 & 0 & 4 & 0 & 1 & 5 & 0 & 0 & 5\\
    \bottomrule
    \end{tabular}}
    \captionof{table}{Words, POS, personalization, and incoming edges for the example sentence.}
    \label{tbl:sentence}
\end{figure}

\begin{figure}
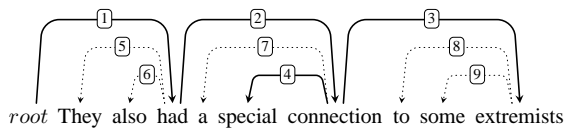

    \centering
    
    \resizebox{\columnwidth}{!}{
      \begin{dependency}
      \begin{deptext}
      $root$ \& They \& also \&  had \&  a \&  special \&  connection \&  to \&  some \&  extremists \\
      \end{deptext}
      \depedge[edge style={black,thick}]{1}{4}{1}
      \depedge[edge style={black,thick}]{4}{7}{2}
      \depedge[edge style={black,thick}]{7}{10}{3}
      \depedge[edge style={black,thick}]{7}{6}{4}
      \depedge[edge style={black,dotted}]{4}{2}{5}
      \depedge[edge style={black,dotted}]{4}{3}{6}
      \depedge[edge style={black,dotted}]{7}{5}{7}
      \depedge[edge style={black,dotted}]{10}{8}{8}
      \depedge[edge style={black,dotted}]{10}{9}{9}
      \end{dependency}}
      \captionof{figure}{Example dependency tree predicted by the algorithm.  \label{fig:tree}}

   \end{figure}



Table \ref{tbl:graphmatrix} shows the directed multigraph used for PR in detail. We can see, e.g., that the four incoming edges for the verb ``had'' from the two nouns, plus from the adverb ``also'' and the pronoun ``They''. 




After running PR, we obtain the following ranking for content words:\\
$C = \langle$had,connection,extremists,special$\rangle$\\
Even though the verb has four incoming edges and the nouns have five each, the personalization makes the verb the highest-ranked word.

\subsection{Decoding}
Once $C$ is calculated, we can follow the algorithm in Fig.~\ref{fig:alg} to obtain a dependency parse. Table \ref{tab:trace} shows a trace of the algorithm, with $C = \langle$had,connection,extremists,special$\rangle$ and $F = \{$They,also,a,to,some\}.

\begin{figure}[ht]
    \resizebox{\columnwidth}{!}{
    \begin{tabular}{ c c c c  }
    \toprule
    $it$ & $word$ & $h$ & $H$ \\
    1 & had & $root$& $\emptyset$ \\
    2 & connection & $had$& $\{had\}$\\
    3 & extremists & $had$& $\{had,connection\}$\\
    4 & special & connection& $\{had,connection,extremists\}$ \\
    \midrule
    5 & They & had& $\{had,connection,extremists,special\}$ \\
    6 & also & had& ... \\
    7 & a & connection& ... \\
    8 & to & extremists &... \\
    9 & some & extremists& ... \\
    \bottomrule
    \end{tabular}}
    \captionof{table}{Algorithm trace for example sentence. $it$: iteration number, $word$: current word, $H$: set of possible heads. \label{tab:trace}}

\end{figure}

The first four iterations calculate the head of content words following their PR, and the following iterations attach the function words in $F$. Finally, Fig. \ref{fig:tree} shows the resulting dependency tree. Full lines are assigned in the first block (content dependents), dotted lines are assigned in the second block (function dependents). The edge labels indicate in which iteration the algorithm has assigned each dependency. Note that the algorithm is deterministic for a certain input POS sequence. Any 10-token sentence with the POS labels shown in Table \ref{tbl:sentence} would yield the same dependency tree.\footnote{The resulting trees always pass the validation script in \url{github.com/UniversalDependencies/tools}.}

\begin{figure}
    \resizebox{\columnwidth}{!}{
    \begin{tabular}{ l c c c c c c c c c }
    \toprule
    $\longrightarrow$ &They & also & had & a & special & connection & to & some & extremists \\
    \midrule
    They & - & & $\bullet$ & & & & $\bullet$ & & \\
    also & & - & $\bullet$& &$\bullet$ & & & & \\
    had & & &- & & & & & & \\
    a & & & & -& &$\bullet$ & & & $\bullet$\\
  special & & & & &- &$\bullet$ & & &$\bullet$ \\
  connection & & &$\bullet$ & &&- & & &$\bullet$ \\
  to & & & & & &$\bullet$ &-& & $\bullet$\\
  some & & & & & &$\bullet$ & &-&$\bullet$ \\
  extremists & & &$\bullet$ & & &$\bullet$ & & &-\\
    \bottomrule
    \end{tabular}}
    \captionof{table}{Matrix representation of the directed graph for the words in the sentence.  }
    \label{tbl:graphmatrix}

\end{figure}

\section{Experiments}
This section describes the data, metrics and comparison systems used to assess the performance of UDP. We evaluate on the test sections of the UD1.2 treebanks \cite{UD12} that contain word forms. If there is more than one treebank per language, we use the treebank that has the canonical language name (e.g., \emph{Finnish} instead of \emph{Finnish-FTB}).  
We use standard unlabeled attachment score (UAS) and evaluate on all sentences of the canonical UD test sets. 

\subsection{Baseline}
\label{sec:baseline}
We compare our UDP system with the performance of a rule-based baseline that uses the head rules in Table~\ref{tbl:results}. The baseline identifies the first verb (or first content word if there are no verbs) as the main predicate, and assigns heads to all words according to the rules in Table~\ref{tab:rules}. We have selected the set of head rules to maximize precision on the development set, and they do not provide full coverage. The system makes any word not covered by the rules (e.g., a word with a POS such as \textsc{x} or \textsc{sym}) either dependent of their left or right neighbor, according to the estimated runtime parameter.

We report the best head direction and its score for each language in Table \ref{tbl:results}. This baseline finds the head of each token based on its closest possible head, or on its immediate left or right neighbor if there is no head rule for the POS at hand, which means that this system does not necessarily yield well-formed tress. Each token receives a head, and while the structures are single-rooted, they are not necessarily connected.
Note that we do not include results for the DMV model by \newcite{klein2004corpus}, as it has been outperformed by a system similar to ours \cite{sogaard2012unsupervised}. 
The usual adjacency baseline for unsupervised dependency parsing, where all words depend on their left or right neighbor, fares much worse than our baseline (20\% UAS  below on average) even with an oracle pick for the best per-language direction, and we do not report those scores.


\subsection{Evaluation setup}
Our system relies solely on POS tags. To estimate the quality degradation of our system under non-gold POS scenarios, we evaluate UDP on two alternative scenarios. The first is predicted POS (UDP$_P$), where we tag the respective test set with TnT \cite{brants2000tnt} trained on each language's training set. The second is a naive type-constrained two-POS tag scenario (UDP$_N$), and approximates a lower bound. We give each word either {\sc content} or {\sc function} tag, depending on the word's frequency. The 100 most frequent words of the input test section receive the {\sc function} tag. 

\noindent
Finally, we compare our parser UDP to a supervised cross-lingual system (MSD). It is a multi-source delexicalized transfer parser, referred to as {\em multi-dir} in the original paper by~\newcite{mcdonald2011multi}. For this baseline we train TurboParser~\cite{martins2013turning} on a delexicalized training set of 20k sentences, sampled uniformly from the UD training data excluding the target language. 
MSD is a competitive and realistic baseline in cross-lingual transfer parsing work. This gives us an indication how our system compares to standard cross-lingual parsers. 


\subsection{Results}
Table~\ref{tbl:results} shows that UDP is a competitive system; because UDP$_G$ is remarkably close to the supervised MSD$_G$ system, with an average difference of 6.4\%. Notably, UDP even outperforms MSD on one language (Hindi). 

\begin{figure}[!htb]
    \centering
 \resizebox{\columnwidth}{!}{
\begin{tabular}{ r l | l l | l l l}
\toprule
Language & BL$_G$ & UDP$_G$ & MSD$_G$ &MSD$_P$   & UDP$_P$ &  UDP$_N$  \\ 
\midrule
Ancient Greek & 42.2 L  & 43.4 & 48.6 & 46.5 &  41.6 & 27.0\\ 
Arabic & 34.8 R & 47.8 & 52.8 & 52.6 & 47.6 & 41.0\\ 
Basque & 47.8 R & 45.0 & 51.2 & 49.3 & 43.1 & 22.8\\ 
Bulgarian & 54.9 R & 70.5 & 78.7 &  76.6 & 68.1 & 27.1\\ 
Church Slavonic & 53.8 L & 59.2 & 61.8 & 59.8 & 59.2 & 35.2\\ 
Croatian & 41.6 L & 56.7 & 69.1 & 65.6 & 54.5 & 25.2\\ 
Czech & 46.5 R & 61.0 & 69.5 & 67.6 & 59.3 & 25.3\\ 
Danish & 47.3 R & 57.9 & 70.2 & 65.6 & 53.8 & 26.9\\ 
Dutch & 36.1 L & 49.5 & 57.0 & 59.2& 50.0 & 24.1\\ 
English & 46.2 R & 53.0 & 62.1 & 59.9 & 51.4 & 27.9\\ 
Estonian & \underline{73.2} R & 70.0 & 73.4 & 66.1 & 65.0 & 25.3\\ 
Finnish & 43.8 R & 45.1 & 52.9 & 50.4 & 43.1 & 21.6\\ 
French & 47.1 R & 64.5 & 72.7 & 70.6 & 62.1 & 36.3\\ 
German & 48.2 R & 60.6 & 66.9 & 62.5 & 57.0 & 24.2\\ 
Gothic & 50.2 L & 57.5 & 61.7 & 59.2 & 55.8 & 34.1\\ 
Greek & 45.7 R & 58.5 & 68.0 & 66.4 & 57.0 & 29.3\\ 
Hebrew & 41.8 R & 55.4 & 62.0 & 58.6 & 52.8 & 35.7\\ 
Hindi & 43.9 R & \textbf{46.3} & 34.6 & 34.5 & \textbf{45.7} & 27.0\\ 
Hungarian & 53.1 R & 56.7 & 58.4 & 56.8 & 54.8 & 22.7\\ 
Indonesian & 44.6 L & 60.6 & 63.6 & 61.0 & 58.4 & 35.3\\ 
Irish & 47.5 R & 56.6 & 62.5 & 61.3 & 53.9 & 35.8\\ 
Italian & 50.6 R & 69.4 & 77.1 & 75.2 & 67.9 & 37.6\\ 
Latin & 49.4 L & 56.2 & 59.8 & 54.9 &52.4 & 37.1\\ 
Norwegian & 49.1 R & 61.7 & 70.8 & 67.3 &58.6 & 29.8\\ 
Persian & 37.8 L & 55.7 & 57.8 & 55.6 & 53.6 & 33.9\\ 
Polish & 60.8 R & 68.4 & 75.6 & 71.7 & 65.7 & 34.6\\ 
Portuguese & 45.8 R & 65.7 & 72.8 & 71.4 & 64.9 & 33.5\\ 
Romanian & 52.7 R & 63.7 & 69.2	 &64.0 & 58.9 & 32.1\\ 
Slovene & 50.6 R & 63.6 & 74.7 & 71.0 & 56.0 & 24.3\\ 
Spanish & 48.2 R & 63.9 & 72.9 & 70.7 & 62.1 & 35.0\\ 
Swedish & 52.4 R & 62.8 & 72.2 & 67.2 & 58.5 & 25.3\\ 
Tamil & \underline{41.4} R & 34.2 & 44.2 & 39.5 & 32.1 & 20.3  \\
\midrule 
\emph{Average} & 47.8	& 57.5 & 63.9 & 61.2 & 55.3&	29.9 \\
\bottomrule
\end{tabular}}
\captionof{table}{UAS for baseline with gold POS (BL$_G$) with direction (L/R) for backoff attachments, UDP with gold POS (UDP$_G$) and predicted POS (UDP$_P$), PR with naive content-function POS (UDP$_N$), and multi-source delexicalized with gold and predicted POS (MSD$_G$ and MSD$_P$, respectively). BL values higher than UDP$_G$ are underlined, and UDP$_G$ values higher than MSD$_G$ are in boldface. }
\label{tbl:results}
\end{figure}

More interestingly, on the evaluation scenario with predicted POS we observe that our system drops only marginally (2.2\%) compared to MSD (2.7\%). In the least robust rule-based setup, the error propagation rate from POS to dependency would be doubled, as either a wrongly tagged head or dependent would break the dependency rules. However, with an average POS accuracy by TnT of 94.1\%, the error propagation is 0.37, i.e, each POS error causes 0.37 additional dependency errors. 
In contrast, for MSD this error propagation is 0.46, thus higher.
\footnote{Err. prop. $=(E(Parse_P)-E(Parse_G))/E(POS_P)$, where $E(x) = 1 - Accuracy(x)$.}

For the extreme POS scenario, content vs. function POS (CF), the drop in performance for UDP is very large, but this might be too  crude an evaluation setup. Nevertheless, UDP, the simple unsupervised system with PageRank, outperforms the adjacency baselines (BL) by $\sim$4\% on average on the two type-based naive POS tag scenario. This difference indicates that even with very deficient POS tags, UDP can provide better structures.

\section{Discussion}
In this section we provide a further error analysis of the UDP parser. We examine the contribution to the overal results of using PageRank to score content words, the behavior of the system across different parts of speech, and we assess the robustness of UDP on  text from different domains.

\subsection{PageRank contribution}
UDP depends on PageRank to score content words, and on two-step decoding to ensure the leaf status of function words. In this section we isolate the constribution of both parts. We do so by comparing the performance of BL, UDP, and UDP$_{NoPR}$, a version of UDP where we disable PR and rank content words according to their reading order, i.e., the first word in the ranking is the first word to be read, regardless of the specific language's script direction. The baseline BL described in \ref{sec:baseline} already ensures function words are leaf nodes, because they have no listed dependent POS in the head rules. The task of the decoding steps is mainly to ensure the resulting structures are well-formed dependency trees.

\noindent 
If we measure the difference between UDP$_{NoPR}$ and BL, we see that UDP$_{NoPR}$ contributes with 4 UAS points on average over the baseline. Nevertheless, the baseline is oracle-informed about the language's best branching direction, a property that UDP does not have. Instead, the decoding step determines head direction as described in Section \ref{sec:directionality}. Complementarily, we can measure the contribution of PR by observing the difference between regular UDP and UDP$_{NoPR}$. The latter scores on average 9 UAS points lower than UDP. These 9 points are caused by the difference attachment of content words in the first decoding step.

\subsection{Breakdown by POS} 
\label{sec:analysis}

UD is a constantly improving effort, and not all v1.2 treebanks have the same level of formalism compliance. Thus, the interpretation of, e.g., the \textsc{aux}--\textsc{verb} or \textsc{det}--\textsc{pron} distinctions might differ across treebanks. However, we ignore these differences in our analysis and consider all treebanks to be equally compliant. 

The root accuracy scores oscillate around an average of 69\%, with 
Arabic and Tamil (26\%) and Estonian (93\%) as outliers. Given the PR personalization (Sec.\ \ref{sec:pagerank}), UDP has a strong bias for choosing the first verb as main predicate. Without personalization, performance drops 2\% on average. This difference is consistent even for verb-final languages like Hindi, given that the main verb of a simple clause will be its only verb, regardless of where it appears. Moreover, using PR personalization makes the ranking calculations converge a whole order of magnitude faster. 

The bigram heuristic to determine adposition direction succeeds at identifying the predominant pre- or postposition preference for all languages (average \textsc{adp} UAS of 75\%). The fixed direction for the other functional POS is largely effective, with few exceptions, e.g., \textsc{det} is consistently right-attaching on all treebanks except Basque (average overall \textsc{det} UAS of 84\%, 32\% for Basque). These alternations could also be estimated from the data in a manner similar to \textsc{adp}.
Our rules do not make nouns eligible heads for verbs. As a result, the system cannot infer relative clauses. We have excluded the \textsc{noun} $\rightarrow$ \textsc{verb} rule during development because it makes the hierarchy between verbs and nouns less conclusive.

We have not excluded punctuation from the evaluation. Indeed, the UAS for the \textsc{punct} is low (an average of 21\%, standard deviation of 9.6), even lower than the otherwise problematic \textsc{conj}. Even though conjunctions are pervasive and identifying their scope is one of the usual challenges for parsers, the average UAS for \textsc{conj} is much larger (an average of 38\%, standard deviation of 13.5) than for \textsc{punct}. Both POS show large standard deviations, which indicates great variability. This variability can be caused by linguistic properties of the languages or evaluation datasets, but also by differences in annotation convention.

\subsection{Cross-domain consistency}

Models with fewer parameters are less likely to overfit for a certain dataset. In our case, a system with few, general rules is less likely to make attachment decisions that are very particular of a certain language or dataset. \newcite{plank-vannoord:2010} have shown that rule-based parsers can be more stable to domain shift. We explore if their finding holds for UDP as well, by testing on i) the UD development data as a readily available proxy for domain shift, and ii) manually curated domain splits of select UD test sets.

\begin{figure}[ht]   
     \hfill
\resizebox{\columnwidth}{!}{
\begin{tabular}{ r l l l l l l}

\toprule
 Language & Domain & BL$_G$ & MSD$_G$ & UDP$_G$ & MSD$_P$ & UDP$_P$ \\ 
 \midrule
Bulgarian & \texttt{bulletin} & 48.3 & \underline{67.5} & \underline{67.4} & \underline{65.4}
 & \underline{61.5} \\ 
 & \texttt{legal} & \underline{47.9} & 76.9 & 69.2 & 73.0 & 68.6 \\ 
 & \texttt{literature} & 53.6 & 74.2 & 69.0 & 72.8 & 66.6 \\ 
 & \texttt{news} & 49.3 & 74.6 & 70.2 & 73.0 & 68.2 \\ 
 & \texttt{various} & 51.4 & 74.2 & 72.5 & 72.6 & 69.5 \\ 
\midrule
Croatian & \texttt{news} & \underline{41.2} & \underline{62.4} & 57.9 & 61.8 & \underline{52.2} \\ 
 & \texttt{wiki} & 41.9 & 64.8 & \underline{55.8} & \underline{58.2} & 56.3 \\ 
\midrule
English & \texttt{answers} & 44.1 & 61.6 & 55.9 & 59.5 & 53.7 \\ 
 & \texttt{email} & 42.8 & 58.8 & 52.1 & 57.1 & 56.3 \\ 
 & \texttt{newsgroup} & \underline{41.7} & 55.5 & \underline{49.7} & 52.9 & 51.1 \\ 
 & \texttt{reviews} & 47.4 & 66.8 & 54.9 & 63.9 & 52.2 \\ 
 & \texttt{weblog}  & 43.3 & \underline{51.6} & 50.9 & \underline{49.8} & 53.8 \\ 
 & \texttt{magazine}$\dagger$ & 41.4 & 60.9 & 55.6 & 58.4 & 53.3 \\ 
 & \texttt{bible}$\dagger$ & 38.4 & 56.2 & 56.2 & 56.8 & 48.6 \\ 
 & \texttt{questions}$\dagger$ & 38.7 & 69.7 & 55.6 & 60.5 & \underline{47.2} \\ 
\midrule
Italian & \texttt{europarl} & 50.8 & \underline{64.1} & \textbf{70.6} & \underline{62.7} & \textbf{69.7} \\ 
 & \texttt{legal} & 51.1 & 67.9 & \textbf{69.0} & \underline{64.4} & \textbf{67.2} \\ 
 & \texttt{news} & 49.4 & 68.9 & 67.5 & 67.0 & \underline{65.3} \\ 
 & \texttt{questions} & \underline{48.7} & 80.0 & 77.0 & 79.1 & 76.1 \\ 
 & \texttt{various} & 49.7 & 67.8 & \textbf{69.0} & 65.3 & \textbf{67.6} \\ 
 & \texttt{wiki} & 51.8 & 71.2 & 68.1 & 70.3 & 66.6 \\
\midrule
Serbian & \texttt{news} & 42.8 & \underline{68.0} & 58.8 & 65.6 & \underline{53.3} \\ 
 & \texttt{wiki} & \underline{42.4} & 68.9 & 58.8 & \underline{62.8} & 55.8 \\ 
\bottomrule
\end{tabular}}
\captionof{table}{Evaluation across domains. UAS for baseline with gold POS (BL$_G$), UDP with gold POS (UDP$_G$) and predicted POS (UDP$_P$), and multi-source delexicalized with gold and predicted POS (MSD$_G$ and MSD$_P$). English datasets marked with $\dagger$ are in-house annotated. Lowest results per language underlined. Bold: UDP outperforms MSD. }
\label{tbl:domains}
\end{figure}

\noindent\textbf{Development sets  } We have used the English development data to choose which relations would be included as head rules in the final system (Table \ref{tab:rules}). It would be possible that some of the rules are indeed more befitting for the English data or for that particular section. 

However, if we regard the results for UDP$_G$ in Table \ref{tbl:results}, we can see that there are 24 languages (out of 32) for which the parser performs better than for English. This result indicates that the head rules are general enough to provide reasonable parses for languages other than the one chosen for development. If we run UDP$_G$ on the development sections for the other languages, we find the results are very consistent. Any language scores on average $\pm 1$ UAS with regards to the test section. There is no clear tendency for either section being easier to parse with UDP.

\noindent\textbf{Cross-domain test sets} To further assess the cross-domain robustness, we retrieved the domain (genre) splits from the test sections of the UD treebanks where the domain information is available as sentence metadata: from Bulgarian, Croatian, and Italian. We also include a UD-compliant Serbian dataset which is not included in the UD release but which is based on the same parallel corpus as Croatian and has the same domain splits \cite{agic2015universal}. When averaging we pool Croatian and Serbian together as they come from the same dataset. 

For English, we have obtained the test data splits matching the sentences from the original distribution of the English Web Treebank. In addition to these already available datasets, we have annotated three different datasets to assess domain variation more extensively, namely the first 50 verses of the King James Bible, 50 sentences from a magazine, and 75 sentences from the test split in QuestionBank \cite{judge2006questionbank}. We include the third dataset to evaluate strictly on questions, which we could do already in Italian. While the \texttt{answers} domain in English is made up of text from the Yahoo! Answers forum, only one fourth of the sentences are questions. Note these three small datasets are not included in the results on the canonical test sections in Table \ref{tbl:results}.
\begin{figure}
\resizebox{\columnwidth}{!}{
\begin{tabular}{ r  r r r r r }
\toprule
 Language  & BL$_G$ & MSD$_G$ & UDP$_G$ & MSD$_P$ & UDP$_P$ \\ 
 \midrule
Bulgarian & 50.1$\pm$2.4 & 73.5$\pm$3.5 & 69.7$\pm$1.8 & 71.3$\pm$3.3 & 66.9$\pm$3.2\\ 
Croatian+Serbian & 42.1$\pm$0.7 & 66.0$\pm$3.0 & 57.8$\pm$1.4 & 62.1$\pm$3.0 & 54.4$\pm$2.0\\ 
English & 42.2$\pm$2.8 & 60.1$\pm$6.2 & 53.9$\pm$2.5 & 57.3$\pm$4.3 & 52.0$\pm$3.3\\ 
Italian & 50.3$\pm$1.2 & 70.0$\pm$5.4 & 70.1$\pm$3.3 & 68.1$\pm$6.0 & 68.7$\pm$3.9\\ 
\midrule
\textit{Average Std.} & 1.8 & 4.5 & 2.5 & 4.2 & 3.1\\ 
\bottomrule
\end{tabular}}
\captionof{table}{\label{tbl:averagedomains} Average language-wise domain evaluation. We report average UAS and standard deviation per language. The bottom row provides the average standard deviation for each system.  }
\label{tbl:results-mean-stdev}
\end{figure}

Table \ref{tbl:results-mean-stdev} summarizes the per-language average score and standard deviation, as well as the macro-averaged standard deviation across languages. UDP has a much lower standard deviation across domains compared to MSD. This holds across languages. We attribute this higher stability to UDP being developed to satisfy a set of general properties of the UD syntactic formalism, instead of being a data-driven method more sensitive to sampling bias. This holds for both the gold-POS and predicted-POS setup. The differences in standard deviation are unsurprisingly smaller in the predicted POS setup. In general, the rule-based UPD is less sensitive to domain shifts than the data-driven MSD counterpart, confirming earlier findings~\cite{plank-vannoord:2010}.

Table \ref{tbl:domains} gives the detailed scores per language and domain. From the scores we can see that presidential \texttt{bulletin}, \texttt{legal} and \texttt{weblogs} are amongst the hardest domains to parse. However, the systems often do not agree on which domain is hardest, with the exception of Bulgarian \texttt{bulletin}. 
Interestingly, for the Italian data and some of the hardest domains UDP outperforms MSD, confirming that it is a robust baseline.

\subsection{Comparison to full supervision}
In order to assess how much information the simple principles in UDP provide, we measure how many gold-annotated sentences are necessary to reach its performance, that is, after which size the treebank provides enough information for training that goes beyond the simple linguistic principles outlined in Section \ref{sec:method}.

For this comparison we use a first-order non-projective TurboParser \cite{martins2013turning} following the setup of~\newcite{agic2016multilingual}. The supervised parsers require around 100 sentences to reach UDP-comparable performance, namely a mean of 300 sentences and a median of 100 sentences, with Bulgarian (3k), Czech (1k), and German (1.5k) as outliers. The difference between mean and median shows there is great variance, while UDP provides very constant results, also in terms of POS and domain variation.

\section{Conclusion}

We have presented UDP, an unsupervised dependency parser for Universal Dependencies (UD) that makes use of personalized PageRank and a small set of head-dependent rules. The parser requires no training data and estimates adposition direction directly from the input. 

We achieve competitive performance on all but two UD languages, and even beat a multi-source delexicalized parser (MSD) on Hindi. We evaluated the parser on three POS setups and across domains. Our results show that UDP is less affected by deteriorating POS tags than MSD, and is more resilient to domain changes. Given how much of the overall dependency structure can be explained by this fairly system, we propose UDP as an additional UD parsing baseline. The parser, the in-house annotated test sets, and the domain data splits are made freely available.\footnote{\url{https://github.com/hectormartinez/ud_unsup_parser}}

UD is a running project, and the guidelines are bound to evolve overtime. Indeed, the UD 2.0 guidelines have been recently released. UDP can be augmented with edge labeling for some deterministic labels like \texttt{case} or \texttt{det}.
Some further constrains can be incorporated in UDP. Moreover, the parser makes no special treatment of multiword expression that would require a lexicon, coordinations or proper names. All these three kinds of structures have a flat tree where all words depend on the leftmost one. While coordination attachment is a classical problem in parsing and out of the scope of our work, a proper name sequence can be straightforwardly identified from the part-of-speech tags, and it falls thus in the area of structures predictable using simple heuristics.  Moreover, our use of PageRank could be expanded to directly score the potential dependency edges instead of words, e.g., by means of edge reification.

\section*{Acknowledgments} We thank the anonymous reviewers for their valuable feedback. H\'{e}ctor Mart\'{i}nez Alonso is funded by the French DGA project VerDi. Barbara Plank thanks the Center for Information Technology of the University of Groningen for the HPC cluster. Željko Agić and Barbara Plank thank the Nvidia Corporation for supporting their research. Anders Søgaard is funded by the ERC Starting Grant LOWLANDS No. 313695.

\bibliography{upud}
\bibliographystyle{eacl2017}

\end{document}